\documentclass{article}
\usepackage{spconf,amsmath}

\usepackage{graphicx}
\usepackage{amssymb}
\usepackage{bm}
\usepackage{caption}
\usepackage{scalefnt}
\usepackage{multirow}
\usepackage{here}
\usepackage{siunitx}
\usepackage{url}
\usepackage[breaklinks]{hyperref}
\usepackage{breakurl}
\usepackage{color}

\title{Multilingual End-to-End Speech Translation}
\name{Hirofumi Inaguma$^1$, Kevin Duh$^2$, Tatsuya Kawahara$^1$, Shinji Watanabe$^2$}
\address{$^1$Graduate School of Informatics, Kyoto University, Kyoto, Japan\\
$^2$Center for Language and Speech Processing, Johns Hopkins University, Baltimore, MD, USA
}

\begin{document}
\ninept
\maketitle
\begin{abstract}
In this paper, we propose a simple yet effective framework for multilingual end-to-end speech translation (ST), in which speech utterances in source languages are directly translated to the desired target languages with a universal sequence-to-sequence architecture.
While multilingual models have shown to be useful for automatic speech recognition (ASR) and machine translation (MT), this is the first time they are applied to the end-to-end ST problem.
We show the effectiveness of multilingual end-to-end ST in two scenarios: {\em one-to-many} and {\em many-to-many} translations with publicly available data.
We experimentally confirm that multilingual end-to-end ST models significantly outperform bilingual ones in both scenarios.
The generalization of multilingual training is also evaluated in a transfer learning scenario to a very low-resource language pair.
All of our codes and the database are publicly available to encourage further research in this emergent multilingual ST topic\footnote{Available at \url{https://github.com/espnet/espnet}.}.
\end{abstract}
%
\section{Introduction}\label{sec:intro}
Breaking the language barrier for communication is one of the most attractive goals.
For several decades, the speech translation (ST) task has been designed by processing speech with automatic speech recognition (ASR), text normalization (e.g. punctuation restoration, case normalization etc.), and machine translation (MT) components in a cascading manner \cite{ney1999speech,punctuation_insertion}.
Recently, end-to-end speech translation (E2E-ST) with a sequence-to-sequence model has attracted attention for its extremely simplified architecture without complicated pipeline systems \cite{listen_and_translate,google_fisher_st,audiobook_st}.
By directly translating speech signals in a source language to text in a target language, the model is able to avoid error propagation from the ASR module, and also leverages acoustic clues in the source language, which have shown to be useful for translation \cite{acoustic_for_nmt}.
Moreover, it is more memory- and computationally efficient since complicated decoding for the ASR module and the latency occurring between ASR and MT modules can be bypassed.

Although end-to-end optimization demonstrates competitive results compared to traditional pipeline systems \cite{audiobook_st, st_distillation} and even outperforms them in some corpora \cite{google_fisher_st,google_tts_augmentation}, these models are usually trained with a single language pair only (i.e. bilingual translation).
There is a realistic scenario in the applications of ST models when a speech utterance is translated to multiple target languages in a lecture, news reading, and conversation domains.
For example, TED talks are mostly conducted in English and translated to more than 70 languages in the official website \cite{wit3}.
In these cases, it is a natural choice to support translation of multiple language pairs from speech.

A practical approach for multilingual ST is to construct (mono- or multi-lingual) ASR and (bi- or multi-lingual) MT systems separately and combine them as in the conventional pipeline system \cite{dessloch2018kit}.
Thanks to recent advances in sequence-to-sequence modeling, we can build strong multilingual ASR \cite{watanabe2017language,toshniwal2018multilingual,dalmia2018sequence,jj_slt18,inaguma2019transfer}, and MT systems \cite{multi_nmt_google,multi_nmt_kit_iwslt16,multi_nmt_kit_iwslt17} even with a single model.
However, when speech utterances come from multiple languages, mis-identification of the the source language by the ASR system disables the subsequent MT system from translating properly since it is trained to consume text in the correct source language\footnote{In case of \textit{one-to-many} situation, this does not occur since only the monolingual ASR is required. However, error propagation from the ASR module and latency between the ASR and MT modules is still problematic.}.
In addition, text normalization, especially punctuation restoration, must be conducted for ASR outputs in each source language, from which additional errors could be propagated.

\begin{figure}[!t]
    \centering
    \includegraphics[width=0.95\linewidth]{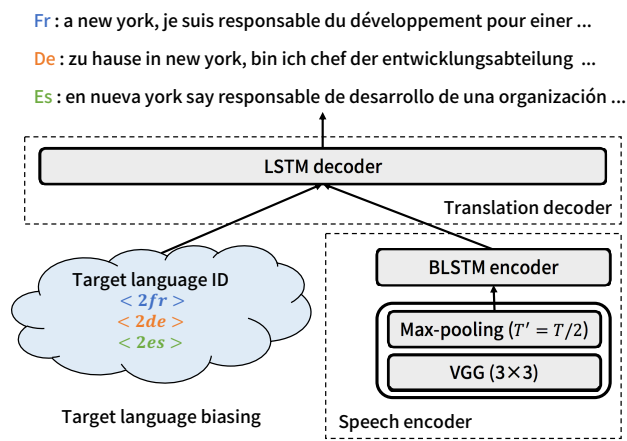}
    \vspace{1mm}
    \caption{System overview for the multilingual end-to-end speech translation model}
    \label{fig:system_overview}
    \vspace{-3mm}
\end{figure}

In this paper, we propose a simple and effective approach to perform multilingual E2E-ST by leveraging a universal sequence-to-sequence model (see Figure \ref{fig:system_overview}).
Our framework is inspired by \cite{multi_nmt_google}, where all parameters are shared among all language pairs, which also enables zero-shot translations.
By building the multilingual E2E-ST system with a universal architecture, it is free from the source language identification and the complexities of training and decoding pipelines are drastically reduced.
Furthermore, we do not have to care about which parameters to share among multiple language pairs, which can be learned automatically from training data.
To the best of our knowledge, this is the first attempt to investigate multilingual training for the E2E-ST task.

We conduct experimental evaluations with three publicly available corpora: Fisher-CallHome Spanish (Es$\to$En) \cite{fisher_callHome}, Librispeech (En$\to$Fr) \cite{librispeech_french}, and Speech-Translation TED corpus (En$\to$De) \cite{jan2018iwslt}.
We evaluate \textit{one-to-many} (O2M) and \textit{many-to-many} (M2M) translations by combining these corpora and confirm significant improvements by multilingual training in both scenarios.
Next, we evaluate the generalization of multilingual E2E-ST models by performing transfer learning to a very low-resource ST task: Mboshi (Bantu C25)$\to$Fr corpus (4.4 hours) \cite{mboshi_french}.
We show that multilingual pre-training of the seed E2E-ST models improves the performance in the low-resource language pair unseen during training, compared to bilingual pre-training.
Our codes are put to the public project so that results can be reproducible and strictly compared in the same pre-processing (e.g., data split, text normalization, and feature extraction etc.), model implementation, and evaluation pipelines.


\section{Background: Speech Translation}\label{sec:model}
In this section, we describe the architecture of the pipeline and end-to-end speech translation (ST) system.
Our ASR, MT, and ST systems are all based on attention-based RNN encoder-decoder models\footnote{We leave to investigate Transformer architectures \cite{transformer} for future work. However, our framework is model agnostic and can be applied to any sequence-to-sequence models.} \cite{attention_nmt_bahdanau,las}.
Let ${\bm x}^{\rm src}$ be the input speech features in a source language, ${\bm y}^{\rm src}$ and ${\bm y}^{\rm tgt}$ be the corresponding reference transcription and translation, respectively.
In this work, we adopt a character-level unit both for source and target references\footnote{Although we also conducted experiments with byte-pair-encoding (BPE) \cite{sennrich2015neural}, the character unit is better than BPE in all settings due to the data sparseness issue. Therefore, we only report results on the character-level unit.}.

\subsection{Pipeline speech translation}\label{ssec:pipeline_st}
The pipeline ST model is composed of three modules: automatic speech recognition (ASR), text normalization, and neural machine translation (NMT) models \cite{punctuation_insertion}.

\subsubsection{Automatic speech recognition (ASR)}\label{sssec:asr}
We build the ASR module based on hybrid CTC/attention framework \cite{hori2017joint,hybrid_ctc_attention}, where the attention-based encoder-decoder is enforced to learn monotonic alignments by jointly optimizing with Connectionist Temporal Classification (CTC) objective function \cite{ctc_graves}.
Our ASR model consists of three modules: the speech encoder, transcription decoder, and the softmax layer for calculating the CTC loss.
The speech encoder transforms input speech features ${\bm x}^{\rm src}$ into a high-level continuous representation, and then the transcription decoder generates a probability distribution $P_{\rm asr}({\bm y^{\rm src}}|{\bm x}^{\rm src})=\prod_{i}{P_{\rm asr}({y^{\rm src}_{i}}|y^{\rm src}_{<i},{\bm x}^{\rm src})}$ conditioned over all previously generated tokens.
We adopt a location-based scoring function \cite{attention_nips2015}.
During training, parameters are updated so as to minimize the linear interpolation of the negative log-likelihood $\mathcal{L}_{\rm att}=-\log P_{\rm att}({\bm y^{\rm src}}|{\bm x}^{\rm src})$ and the CTC loss
$\mathcal{L}_{\rm ctc}=-\log P_{\rm ctc}({\bm y^{\rm src}}|{\bm x}^{\rm src})$ with a tunable parameter $\lambda\ (0 \leq \lambda \leq 1)$: $\mathcal{L}_{\rm asr}= (1 - \lambda) \mathcal{L}_{\rm att} + \lambda \mathcal{L}_{\rm ctc}$.
During the inference, left-to-right beam search decoding is performed jointly with scores from both an external recurrent neural network language model (RNNLM) \cite{rnnlm} (referred to as \textit{shallow fusion}) and the CTC outputs.
We refer the readers to \cite{hori2017joint,hybrid_ctc_attention} for more details.

For multilingual ASR models, we prepend the corresponding language ID to reference labels so that the decoder can jointly identify the target language while recognizing speech explicitly, which can be regarded as multi-task learning with ASR and language identification tasks \cite{watanabe2017language}.

\vspace{-1mm}
\subsubsection{Text normalization}\label{ssec:text_normalization}
In this work, we skip punctuation restoration for the simplicity\footnote{In this paper, we use lowercased references. Therefore, we do not consider truecasing as text normalization.}.
Instead, we train the MT model so that it translates source references without punctuation marks to target references with them, where text normalization task is jointly conducted with the MT task and it can be seen as multi-task learning.
During inference, the MT model consumes hypotheses from the ASR model.

\vspace{-1mm}
\subsubsection{Neural machine translation (NMT)}\label{sssec:nmt}
Our NMT model consists of the source embedding, text encoder, and translation decoder.
The text encoder maps a sequence of source tokens ${\bm y}^{\rm src}$ into the distributed representation following the source embedding layer.
The translation decoder generates a probability distribution $P({\bm y^{\rm tgt}}|{\bm y^{\rm src}})$.
The only differences between the transcription and translation decoders are the score function for the attention mechanism.
We adopt an additive scoring function \cite{attention_nmt_bahdanau}.
Optimization is performed so as to minimize the negative log-likelihood $- \log P({\bm y^{\rm tgt}}|{\bm y}^{\rm src})$.


\vspace{-1mm}
\subsection{End-to-end speech translation (E2E-ST)}\label{ssec:e2e_st}
Our end-to-end speech translation (E2E-ST) model is composed of the speech encoder and translation decoder.
To compare strictly, we use the same speech encoder and translation decoder as ASR and NMT tasks, respectively.
Parameters are updated so as to minimize the negative log-likelihood $- \log P({\bm y^{\rm tgt}}|{\bm x}^{\rm src})$.



\begin{table*}[t]
  \centering
  \begingroup
	\footnotesize
        \scalebox{1.0}[1.0]{
    \begin{tabular}{l|l|c|c|c|c|c} \hline
    Translation & Corpus & {\bf \em \#hours} & {\bf \em \#utterances} & {\bf \em \#words} & {\bf \em \#vocab} & {\bf \em domain} \\ \hline
       \multirow{3}{*}{Bilingual}
      & (A) Fisher-CallHome Spanish (Es$\to$En) & 170 & 138 k & 1.7 M & 66 & conversation \\
      & (B) Librispeech (En$\to$Fr) & 99 & 45 k$\dag$ & 0.8 M & 112 & reading \\
      & (C) ST-TED (En$\to$De) & 203 & 133 k & 2.2 M & 109 & lecture \\ \hline
      One-to-many (O2M) & (B) + (C)  (En$\to$\{Fr, De\}) & 302 & 178 k & 3.3 M & 153 & mixed \\ \hline
      Many-to-many (M2Ma) & (A) + (B) (\{En, Es\}$\to$\{Fr, En\}) & 269 & 183 k & 2.8 M & 121 & mixed \\
      Many-to-many (M2Mb) & (A) + (C) (\{En, Es\}$\to$\{De, En\}) & 373 & 272 k & 4.0 M & 119 & mixed \\
      Many-to-many (M2Mc) & (A) + (B) + (C) (\{En, Es\}$\to$\{Fr, De, En\}) & 472 & 317 k & 5.1 M & 157 & mixed \\ \hline
    \end{tabular}
    }
    \vspace{-2mm}
    \caption{Statistics in each corpus. Each value is calculated \underline{after} normalizing references and removing short and long utterances. \textit{Speed perturbation} based data augmentation \cite{speed_perturbation} is not performed here. $\dag$Two translation references are prepared per source speech utterance.}\label{tab:data}
    \vspace{-4mm}
  \endgroup
\end{table*}

\section{Multilingual E2E speech translation}\label{sec:multi_st}
We now propose an efficient framework that extends the bilingual E2E-ST model described previously to a multilingual one.

\subsection{Universal sequence-to-sequence model}
We adopt a universal sequence-to-sequence architecture instead of preparing separate parameters per language pair for four reasons.
First, E2E-ST can be generally considered as a more challenging task than MT due to its more complex encoder, which requires more parameters (e.g., VGG+BLSTM).
In addition, training sentences in standard ST corpora are much smaller than MT tasks ($<$300k) although input speech frames are much longer than text.
Therefore, by sharing all parts, the total number of parameters are also reduced considerably and the E2E-ST model can have more training samples for better translation performance.
Furthermore, it is not necessary to change the existing architecture.
Second, we do not have to carefully pre-define a mini-batch scheduler for the language cycle as in \cite{multi_nmt_multiway} (see Section \ref{ssec:mixed_data_training}).
Third, translation performance in low-resource directions can be improved by the aid of high-resource language pairs.
Fourth, we can realize zero-shot translation in a direction which has never been seen during training \cite{multi_nmt_google}.

\subsection{Target language biasing}\label{ssec:target_language_biasing}
To perform translations for multiple target languages with a single decoder, we have to specify a target language to translate to.
In \cite{multi_nmt_google,multi_nmt_kit_iwslt16,multi_nmt_kit_iwslt17}, an artificial token to represent the target language  (target language ID) is prepended in the source sentence.
However, this is not suitable for the ST task since the ST encoder directly consumes speech features.
Instead, we replace a start-of-sentence ($\langle sos \rangle$) token in the decoder with a target language ID $\langle 2lang \rangle$ (see Figure \ref{fig:system_overview}).
For example, when English speech is translated to French text, $\langle sos \rangle$ is replaced with French ID token $\langle 2fr \rangle$.

\vspace{-1mm}
\subsection{Mixed data training}\label{ssec:mixed_data_training}
We train multilingual models with mixed training data from multiple languages.
Thus, each mini-batch may contain utterances from different language pairs.
We bucket all samples so that each mini-batch contains utterances of speech frames of the similar lengths regardless of language pairs.
As a result, we can use the same training scheme as the conventional ASR and bilingual ST tasks.

\section{Data}\label{ssec:data}
We build our systems on three speech translation corpora: Fisher-CallHome Spanish, Librispeech, and Speech-Translation TED (ST-TED) corpus.
To the best of our knowledge, these are the only public available corpora recorded with a reasonable size of real speech data\footnote{We noticed publicly available \textit{one-to-many} multilingual ST corpus \cite{mustc} right before submission. However, this dataset has English speech only.}.
The data statistics are summarized in Table \ref{tab:data}.

\subsection{Bilingual translation}
\subsubsection*{(A) Fisher-CallHome Spanish: Es$\to$En}
This corpus contains about 170-hours of Spanish conversational telephone speech, the corresponding transcription, and the English translations\footnote{\url{https://github.com/joshua-decoder/Fisher-CallHome-corpus}} \cite{fisher_callHome}.
Following \cite{google_fisher_st,fisher_callHome,some_insights}, we report results on the five evaluation sets: {\em dev}, {\em dev2}, and {\em test} in Fisher corpus (with four references), and {\em devtest} and {\em evltest} in CallHome corpus (with a single reference).
We use the Fisher/\textit{train} as the training set and Fisher/\textit{dev} as the validation set.
All punctuation marks except for apostrophe are removed during evaluation in ST and MT tasks to compare with previous works \cite{google_fisher_st,fisher_callHome}.

\subsubsection*{(B) Librispeech: En$\to$Fr}
This corpus is a subset of the original Librispeech corpus \cite{librispeech} and contains 236-hours of English read speech, the corresponding transcription, and the French translations \cite{librispeech_french}.
We use the clean 99-hours of speech data for the training set \cite{audiobook_st}.
Translation references in the training set are augmented with Google Translate following \cite{audiobook_st}, so we have two French references per utterance.
We use the {\em dev} set as the validation set and report results on the {\em test} set.

\subsubsection*{(C) Speech-Translation TED (ST-TED): En$\to$De}
This data contains 271-hours of English lecture speech, the corresponding transcription, as well as the German translation\footnote{\url{https://sites.google.com/site/iwsltevaluation2018/Lectures-task}}.
Since the original training set includes a lot of noisy utterances due to low alignment quality, we take a data cleaning strategy.
We first force-aligned all training utterances with a Gentle forced aligner\footnote{\url{https://github.com/lowerquality/gentle}} based on Kaldi \cite{kaldi}, then excluded all utterances where all words in the transcription were not perfectly aligned with the corresponding audio signal \cite{iwslt18_data_cleaning}.
This process reduced from 171,121 to 137,660 utterances.
We sampled two sets of 2k utterances from the cleaned training data as the validation and test sets, respectively (totally 4k utterances).
Note that all sets have no text overlap and are disjoint regarding speakers, and data splits are available in our codes.
We report results on this test set and {\em tst2013}.
{\em tst2013} is one of the test sets provided in IWSLT2018 evaluation campaign. Since there are no human-annotated time alignment provided in these test sets, we decided to sample the disjoint test set from the training data with alignment information.

\subsection{Multilingual translation}
We perform experiments in two scenarios: {\em one-to-many} (O2M) and {\em many-to-many} (M2M)\footnote{For {\em many-to-one} (M2O) scenario, none of the corpora combinations exists in publicly available corpora, therefore we leave the exploration of this task for future work. However, O2M and M2M are the realistic scenarios for multilingual speech translation as mentioned in Section \ref{sec:intro}.}.

\subsubsection*{One-to-many (O2M)}
For one-to-many (O2M) translation, speech utterances in a source language are translated to multiple target languages.
We concatenate Librispeech (En$\to$Fr) and ST-TED (En$\to$De), and build models for En$\to$\{Fr, De\} translations (see Table \ref{tab:data}).

\subsubsection*{Many-to-many (M2M)}
For many-to-many (M2M) translation, speech utterances in multiple source languages are translated to all target languages given in training.
We can regard this task as a more challenging optimization problem than O2M and M2O translations.
We concatenate Librispeech (En$\to$Fr) and Fisher-CallHome Spanish (Es$\to$En), then build models for \{En, Es\}$\to$\{Fr, En\} translations (M2Ma)\footnote{Readers might think that this scenario is not suitable for the M2M evaluation since French does not appear in source side as in the multilingual MT task \cite{multi_nmt_google}. However, such public corpora are not currently available.}.
Other combinations such as Fisher-CallHome Spanish and ST-TED (\{En, Es\}$\to$\{De, En\}, M2Mb), and all three directions (\{En, Es\}$\to$\{Fr, De, En\}, M2Mc) are also investigated.

\begin{table*}[htbp]
  \begin{center}
    \centering Bi-$\ast$: Bilingual, Mono-$\ast$: Monolingual, Multi-$\ast$: Multilingual\\
    \vspace{1mm}
    \begin{tabular}{c}
      \begin{minipage}{0.58\hsize}
      \begin{flushleft}
        \begingroup
        \footnotesize
        \scalebox{0.92}[0.92]{
        \begin{tabular}{c|l|c|ccccc} \hline
          \multicolumn{2}{c|}{\multirow{3}{*}{Model}} & \multirow{3}{*}{\shortstack{Multi-\\lingual}} & \multicolumn{3}{c|}{Fisher} & \multicolumn{2}{c}{CallHome}   \\ \cline{4-8}
          \multicolumn{2}{c|}{} &  & dev & dev2 & \multicolumn{1}{c|}{test} & devtest & evltest \\ \cline{4-8}
          \multicolumn{2}{c|}{} &  & \multicolumn{5}{c}{BLEU ($\uparrow$)} \\ \cline{4-8} \hline
          \multirow{6}{*}{MT}
            & Bi-SMT \cite{some_insights} & -- & -- & 65.4 & 62.9 & -- & -- \\
            & Bi-NMT \cite{google_fisher_st} & -- & 58.7 & 59.9 & 57.9 & 28.2 & 27.9 \\
            & Bi-NMT \cite{disfluency_st} & -- & 61.9 & 62.8 & 60.4 & -- & -- \\ \cline{2-8}
            & Bi-NMT & -- & {\bf 60.6} & {\bf 62.0} & {\bf 59.6} & {\bf 29.4} & {\bf 28.9} \\
            & Multi-NMT & M2Ma & 50.2 & 50.6 & 49.5 & 22.8 & 22.8 \\
            & Multi-NMT & M2Mb & 57.4 & 58.3 & 56.7 & 27.9 & 27.7 \\
            & Multi-NMT & M2Mc & 56.7 & 57.5 & 56.2 & 27.8 & 27.7  \\\hline
        \multirow{6}{*}{\shortstack{E2E\\ST}}
            & Bi-ST \cite{google_fisher_st} & -- & 46.5 & 47.3 & 47.3 & 16.4 & 16.6 \\
            & \ + ASR task \cite{google_fisher_st} & -- & 48.3 & 49.1 & 48.7 & 16.8 & 17.4 \\ \cline{2-8}
            & (E-B-1) Bi-ST & -- & 40.4 & 41.4 & 41.5 & 14.1 & 14.2 \\
            & (E-Ma-1) Multi-ST & M2Ma & 41.1 & 41.7 & 41.3 & 15.1 & 15.2 \\
            & (E-Mb-1) Multi-ST & M2Mb & 43.5 & 44.5 & 44.2 & 15.3 & 15.8 \\
            & (E-Mc-1) Multi-ST & M2Mc & {\bf 44.1} & {\bf 45.4} & {\bf 45.2} & {\bf 16.4} & {\bf 16.2} \\ \hline
        \multirow{6}{*}{\shortstack{Pipe\\ST}}
            & Mono-ASR/Bi-SMT \cite{some_insights} & -- & -- & -- & 40.4 & -- & -- \\
            & Mono-ASR/Bi-NMT \cite{google_fisher_st} & -- & 45.1 & 46.1 & 45.5 & 16.2 & 16.6 \\ \cline{2-8}
            & (P-B) Mono-ASR/Bi-NMT & -- & 37.3 & 39.6 & 38.6 & 16.8 & 16.5 \\
            & (P-Ma) Multi-ASR/Bi-NMT & M2Ma & {\bf 37.9} & {\bf 40.3} & {\bf 39.2} & {\bf 17.6} & {\bf 17.2} \\
            & (P-Mb) Multi-ASR/Bi-NMT & M2Mb & 37.6 & 39.6 & 38.9 & 17.0 & 17.0 \\
            & (P-Mc) Multi-ASR/Bi-NMT & M2Mc & 37.6 & 39.7 & 38.5 & 17.0 & 16.9 \\ \hline \hline
        \multicolumn{2}{c|}{Model} &  & \multicolumn{5}{c}{WER ($\downarrow$)} \\ \cline{4-8} \hline
        \multirow{5}{*}{ASR}
            & Mono-ASR \cite{google_fisher_st} & -- & 25.7 & 25.1 &23.2	  & 44.5 & 45.3 \\ \cline{2-8}

            & Mono-ASR (Es) & --        & 26.0 & 25.6 & 23.6 & 45.4 & 45.9 \\
            & Multi-ASR (Es, En) & M2Ma & {\bf 25.6} & {\bf 25.0} & {\bf 22.9} & {\bf 43.5} & 44.5 \\
            & Multi-ASR (Es, En) & M2Mb & 25.9 & 25.2 & 23.3 & 44.2 & 44.7 \\
            & Multi-ASR (Es, En) & M2Mc & 26.0 & 25.4 & 23.6 & 44.5 & {\bf 44.2} \\ \hline
        \end{tabular}
        }
        \vspace{-2mm}
        \caption{Results of MT, ST, and ASR systems on Fisher-CallHome Spanish (Es$\to$En). (E-B-1): Bilingual E2E-ST. (E-Ma/Mb/Mc-1): Proposed \textit{many-to-many} (M2M) E2E-ST. (P-B): Bilingual pipeline-ST. (P-Ma/Mb/Mc): M2M pipeline-ST.}\label{tab:result_fisher}
        \endgroup
      \end{flushleft}
      \end{minipage}

      \begin{minipage}{0.1\hsize}
      \vspace{-2mm}
      \end{minipage}

      \begin{minipage}{0.38\hsize}
      \begin{flushright}
        \vspace{-1.8mm}
        \begingroup
        \scalebox{0.84}[0.84]{
        \begin{tabular}{c|l|c|c} \hline
          \multicolumn{2}{c|}{\multirow{2}{*}{Model}} & \multirow{2}{*}{\shortstack{Multi-\\lingual}} & \multirow{2}{*}{BLEU ($\uparrow$)} \\
          \multicolumn{2}{c|}{} &  &  \\ \hline
       	  \multirow{6}{*}{MT}
                 & Bi-NMT \cite{audiobook_st} & -- & 19.2 \\
                 & Google Translate \cite{audiobook_st} & -- & 22.2 \\ \cline{2-4}
                 & Bi-NMT & -- & {\bf 18.3} \\
                 & Multi-NMT & O2M & 16.2 \\
                 & Multi-NMT & M2Ma & 12.2 \\
                 & Multi-NMT & M2Mc & 14.8 \\ \hline
          \multirow{7}{*}{\shortstack{E2E\\ST}}
                 & Bi-ST \cite{audiobook_st} & -- & 12.9 \\
                 & \ + Pre-training + MTL \cite{audiobook_st} & -- & 13.4 \\
                 & Bi-ST + KD \cite{st_distillation} & -- & 17.0 \\ \cline{2-4}
                 & (E-B-1) Bi-ST & -- & 15.7 \\
                 & (E-O-1) Multi-ST & O2M & 17.2 \\
                 & (E-Ma-1) Multi-ST & M2Ma & 16.4 \\
                 & (E-Mc-1) Multi-ST & M2Mc & {\bf 17.3} \\ \hline
         \multirow{5}{*}{\shortstack{Pipe\\ST}}
                 & Mono-ASR/Bi-NMT \cite{audiobook_st} & -- & 14.6 \\ \cline{2-4}
                 & (P-B) Mono-ASR/Bi-NMT & -- & 15.8  \\
                 & (P-O) Mono-ASR$\dag$/Bi-NMT & O2M & {\bf 16.7} \\
                 & (P-Ma) Muti-ASR/Bi-NMT & M2Ma & 16.4 \\
                 & (P-Mc) Muti-ASR/Bi-NMT & M2Mc & {\bf 16.7} \\ \hline \hline
         \multicolumn{2}{c|}{Model} &  & WER ($\downarrow$) \\ \hline
         \multirow{5}{*}{ASR}
                 & Mono-ASR \cite{audiobook_st} & -- & 17.9 \\ \cline{2-4}
                 & Mono-ASR (En) & -- & 9.0 \\
                 & Mono-ASR$\dag$ (En) & O2M & {\bf 6.6} \\
                 & Multi-ASR (En, Es) & M2Ma & 8.6 \\
                 & Multi-ASR (En, Es) & M2Mc & 6.8 \\ \hline
        \end{tabular}
        }
        \vspace{-2mm}
        \caption{Results of MT, ST, and ASR systems on Librispeech (En$\to$Fr). $\dag$Training data is augmented with ST-TED. (E-O-1): Proposed \textit{one-to-many} (O2M) E2E-ST. (P-O): O2M pipeline-ST.}\label{tab:result_libri_bleu}
        \endgroup
      \end{flushright}
      \end{minipage}

    \end{tabular}
    \vspace{-7mm}
  \end{center}
\end{table*}

\section{Experimental evaluations}\label{sec:exp}
\subsection{Settings}
For data pre-processing of references in all languages, we lowercased and normalized punctuation, followed by tokenization with the {\tt tokenizer.perl} script in the Moses toolkit\footnote{\url{https://github.com/moses-smt/mosesdecoder}}.
For source references, we further removed all punctuation marks except for apostrophe.
We report case-insensitive BLEU \cite{bleu} with the {\tt multi-bleu.perl} script in Moses.
The character vocabulary was created jointly with both source and target languages.

We used 80-channel log-mel filterbank coefficients with 3-dimensional pitch features, computed with a 25ms window size and shifted every 10 ms using Kaldi \cite{kaldi}, resulting 83-dimensional features per frame.
The features were normalized by the mean and the standard deviation for each training set.
We augmented speech data by a factor of 3 by \textit{speed perturbation} \cite{speed_perturbation}.
We removed utterances having more than 3000 frames or more than 400 characters due to the GPU memory efficiency.

The speech encoders in ASR and ST models were composed of two VGG blocks \cite{vgg} followed by 5-layers of 1024-dimensional (per direction) bidirectional long short-term memory (LSTM) \cite{lstm}.
Each VGG-like block composed of 2-layers of CNN having a $3\times3$ filter followed by a max-pooling layer with a stride of $2\times2$, which resulted in 4-fold time reduction.
The text encoders in MT models were composed of 2-layers of 1024-dimensional (per direction) BLSTM.
Both transcription and translation decoders were two layers of unidirectional LSTM with 1024-dimensional memory cells.
The dimensions of the attention layer and embeddings for decoders were set to 1024.
We used 2-layers of LSTM LM with 1024 memory cells for shallow fusion as discussed in Section \ref{sssec:asr}.

Training was performed using Adadelta \cite{adadelta} for sequence-to-sequence models and Adam \cite{adam} for RNNLM.
For regularization, we adopted dropout \cite{zaremba2014recurrent}, label smoothing \cite{label_smoothing}, scheduled sampling \cite{scheduled_sampling}, and weight decay.
Beam search decoding was performed with a beam width of 20 with CTC and LM scores in the ASR task as shown in Section \ref{sssec:asr}, and a beam width of 10 with a length penalty in ST and MT tasks.
Detailed hyperparameter settings during training and decoding are available in our codes.

\subsection{Baseline results: Bilingual systems}\label{ssec:result_bilingual}
First, we evaluate baseline bilingual MT and ST systems.
Bilingual E2E-ST and pipeline-ST models are labeled (E-B-1) and (P-B) in each table, respectively.

\subsubsection*{(A) Fisher-CallHome Spanish: Es$\to$En}\label{sssec:results_fisher}
We present our results on Fisher-CallHome Spanish (hereafter, Fisher-CallHome) in Table \ref{tab:result_fisher}.
ASR and NMT results were competitive to the previous work \cite{google_fisher_st} while the E2E-ST and pipeline-ST models underperformed it.
Note that our translation decoders in E2E-ST and NMT models were trained so as to predict lowercased references with punctuation marks to compare with multilingual models, unlike previous works \cite{google_fisher_st,fisher_callHome,some_insights}, where all punctuation marks except for apostrophe are removed.
For the comparison of our E2E-ST and pipeline-ST models, the baseline bilingual E2E-ST model (E-B-1) outperformed the pipeline-ST model (P-B) in the Fisher sets but underperformed it in the CallHome sets.
To investigate this discrepancy, we evaluated them with a single reference in the Fisher tests, which results in 26.4/28.2/27.7 (Pipe-ST) vs. 23.5/25.2/24.8 (E2E-ST) and the pipeline system was shown to be better.
This is intuitive since the E2E-ST model skipped the ASR decoder, RNNLM in the source language, and MT encoder parts.

In our preliminary experiments, we confirmed the E2E-ST model can outperform the pipeline system by stacking more BLSTM layers on top of the speech encoder to match the number of parameters between them.
Moreover, pre-training the speech encoder and translation decoder with the corresponding ASR encoder and NMT decoder also drastically improved the performances (see Table \ref{tab:result_pretrain} in Section \ref{ssec:pretrain}).
However, it is worth noting that our goal in this paper is to show the effectiveness of multilingual training for E2E-ST models and therefore we will not seek these directions here.

\subsubsection*{(B) Librispeech: En$\to$Fr}\label{sssec:results_librispeech}
Next, results on Librispeech are shown in Table \ref{tab:result_libri_bleu}.
Monolingual ASR, bilingual E2E-ST (E-B-1), and pipeline-ST (P-B) models outperformed the previous work \cite{audiobook_st}.
The baseline bilingual E2E-ST model (E-B-1) showed the competitive performance compared to the pipeline-ST model (P-B).

\subsubsection*{(C) ST-TED: En$\to$De}\label{sssec:results_iwslt}
Results on ST-TED are shown in Table \ref{tab:result_iwslt18}.
Contrary to the above results, there is a large gap between the bilingual E2E-ST (E-B-1) and pipeline-ST (P-B) models in this corpus.

\begin{table}[!t]
    \centering
    \begingroup
    \scalebox{0.95}[0.95]{
    \begin{tabular}{c|l|c|cc} \hline
      \multicolumn{2}{c|}{\multirow{2}{*}{Model}} & \multirow{2}{*}{\shortstack{Multi-\\lingual}} & test & tst2013 \\ \cline{4-5}
      \multicolumn{2}{c|}{} &  & \multicolumn{2}{c}{BLEU ($\uparrow$)} \\ \hline
        \multirow{4}{*}{MT}
          & Bi-NMT & -- & {\bf 23.0} & {\bf 24.9} \\
          & Multi-NMT & O2M & 18.9 & 20.3 \\
          & Multi-NMT & M2Mb & 17.5 & 18.7  \\
          & Multi-NMT & M2Mc & 17.2 & 18.0 \\ \hline
        \multirow{4}{*}{\shortstack{E2E\\ST}}
            & (E-B-1) Bi-ST & -- & 16.0 & 12.5 \\
            & (E-O-1) Multi-ST & O2M & 17.6 & 14.4 \\
            & (E-Mb-1) Multi-ST & M2Mb & 16.7 & 12.9 \\
            & (E-Mc-1) Multi-ST & M2Mc & {\bf 17.7} & {\bf 14.8} \\ \hline
        \multirow{4}{*}{\shortstack{Pipe\\ST}}
            & (P-B) Mono-ASR/Bi-NMT & -- & 18.1 & 13.1 \\
            & (P-O) Mono-ASR$\dag$/Bi-NMT & O2M & {\bf 18.5} & {\bf 14.0} \\
            & (P-Mb) Multi-ASR$\dag$/Bi-NMT & M2Mb & 17.7 & 12.6 \\
            & (P-Mc) Multi-ASR$\dag$/Bi-NMT & M2Mc & 18.1 & 13.3 \\ \hline \hline
      \multicolumn{2}{c|}{Model} &  & \multicolumn{2}{c}{WER ($\downarrow$)} \\ \hline
        \multirow{4}{*}{ASR}
          & Mono-ASR (En) & -- & 20.3 & 36.6 \\
          & Mono-ASR$\dag$ (En) & O2M & {\bf 19.0} & {\bf 33.9} \\
          & Multi-ASR (En, Es) & M2Mb & 20.5 & 38.7  \\
          & Multi-ASR (En, Es) & M2Mc & 20.1 & 36.5 \\ \hline
    \end{tabular}
    }
    \vspace{-2mm}
    \caption{Results of MT, ST, and ASR systems on ST-TED (En$\to$De). $\dag$Training data is augmented with Librispeech.}\label{tab:result_iwslt18}
    \vspace{-3mm}
    \endgroup
\end{table}

\subsection{Main results: Multilingual systems}\label{ssec:result_multilingual}
We now test multilingual models trained in two scenarios: {\em many-to-many} (M2M) and {\em one-to-many} (O2M) translations.

\subsubsection*{Many-to-many (M2M)}
Results of M2M models on Fisher-CallHome, Librispeech, ST-TED are shown at the (*-Ma/Mb/Mc-1) lines in Table \ref{tab:result_fisher}, Table \ref{tab:result_libri_bleu}, and Table \ref{tab:result_iwslt18}, respectively.
Ma, Mb, and Mc represent M2Ma, M2Mb, and M2Mc, respectively (see Table \ref{tab:data}).

In Fisher-CallHome (Table \ref{tab:result_fisher}), our M2M multilingual E2E-ST models (E-Mb/Mc-1) significantly outperformed the bilingual one (E-B-1) while (E-Ma-1) slightly outperformed (E-B-1) except for Fisher/\textit{test}.
Among three M2M E2E-ST models, (E-Mc-1) showed the best performance, from which we can confirm that additional training data from other language pairs is effective.
Multilingual ASR models slightly outperformed the monolingual ASR model.
Performances of the MT models were degraded by multilingual training due to the domain mismatch especially for punctuation marks (see Table \ref{tab:data}).
In contrast, multilingual E2E-ST models were not affected by the domain mismatch issue since they are not conditioned on the source language text, which is one for the advantages of the end-to-end models.

In all pipeline systems in Fisher-CallHome, we used the bilingual MT model since it showed the best performance.
Pipeline systems with the multilingual ASR (P-M$\ast$) were consistently improved even though WER improvements were very small.
Our multilingual E2E-ST models significantly outperformed all the pipeline models in the Fisher sets.

In Librispeech (Table \ref{tab:result_libri_bleu}), all M2M E2E-ST models (E-Ma/Mc-1) outperformed the bilingual one (E-B-1).
Multilingual ASR models also outperformed the monolingual one.
Pipeline systems (P-Ma/Mc) are improved in proportion to the WER improvements.
However, E2E-ST models got more gains from multilingual training.

In ST-TED (Table \ref{tab:result_iwslt18}), we also confirmed the consistent BLEU improvements by the proposed multilingual framework.
The similar trends can be seen as in Fisher-CallHome and Librispeech.

\subsubsection*{One-to-many (O2M)}
Results of O2M models on Librispeech and ST-TED are shown in Table \ref{tab:result_libri_bleu} and Table \ref{tab:result_iwslt18}, respectively.
We also obtained significant improvements of the E2E-ST models from multilingual training as well as in the M2M scenario on both corpora.
Since the amount of additional training data for O2M and M2Mb from ST-TED is 99-hours (+Librispeech) and 170-hours (+Fisher-CallHome), respectively, and the O2M E2E-ST model is better than the M2Mb E2E-ST model in ST-TED (see Table \ref{tab:result_iwslt18}), we can conclude that {\em O2M training is more effective than M2M training} in terms of data efficiency.
However, the combination of all training data (M2Mc) got a further small gain.
We can confirm the effectiveness of O2M training from WER improvements in the ASR task (6.6 vs. 8.6 at the second and third lines from bottom in Table \ref{tab:result_libri_bleu}).
Thus, further additional multilingual training data could lead to the improvement.
Gains from multilingual training were larger in the E2E-ST model (E-O-1) than in the best pipeline model (P-O)\footnote{The best monolingual ASR $\to$ the best bilingual NMT in Table \ref{tab:result_iwslt18}}.
Considering the fact that the O2M NMT model underperformed the bilingual one, O2M multilingual training benefits from not only additional English speech data but also the direct optimization, which is one of our motivations in this work.

\subsubsection*{Pre-training with the ASR encoder}\label{ssec:pretrain}
Finally, we show results of pre-training with the ASR encoder in Table \ref{tab:result_pretrain}.
We observed improvements by pre-training both in bilingual and multilingual cases, similar to \cite{audiobook_st,google_tts_augmentation,pretraining_st}.
Pre-training with the NMT decoder was not necessarily effective.
The best multilingual E2E-ST with pre-training (E-Mc-2) outperformed the corresponding best pipeline system in most test sets.

In summary, the proposed multilingual framework has shown to be effective regardless of the language combination, corpus domain, and data size.
Although it is possible to improve the pipeline systems by carefully designing the source representations between ASR and MT modules (e.g., adding punctuation restoration module), it can be overcome by simply optimizing the direct mapping from source speech to target text with punctuation marks as we have shown.

\begin{table*}[!t]
    \centering
    \begingroup
    \scalebox{0.90}[0.90]{
    \begin{tabular}{c|l|c|ccccc|c|cc} \hline
      \multicolumn{2}{c|}{\multirow{3}{*}{Model}} & \multirow{3}{*}{\shortstack{Multi-\\lingual}} & \multicolumn{8}{c}{BLEU ($\uparrow$)} \\ \cline{4-11}
      \multicolumn{2}{c|}{} & & \multicolumn{5}{c|}{Fisher-CallHome} & Librispeech & \multicolumn{2}{c}{ST-TED} \\ \cline{4-11}
      \multicolumn{2}{c|}{} &  & dev & dev2 & test & devtest & evltest & test & test & tst2013 \\ \hline
        \multirow{5}{*}{E2E-ST}
            & (E-B-1) Bi-ST & -- & 40.4 & 41.4 & 41.5 & 14.1 & 14.2 & 15.7 & 16.0 & 12.5 \\
            & (E-B-2) \ + ASR-PT & -- & 43.5 & 45.1 & 44.7 & 15.6 & 16.4 & 16.3 & 17.1 & 13.1 \\
            & (E-B-3) \ \ + MT-PT & -- & 44.4 & 45.1 & 45.2 & 15.6 & 15.4 & 16.8 & 17.4 & 13.5 \\ \cline{2-11}
            & (E-Mc-1) Multi-ST & M2Mc & 44.1 & 45.4 & 45.2 & 16.4 & 16.2 & 17.3 & 17.7 & {\bf 14.8} \\
            & (E-Mc-2) \ + ASR-PT & M2Mc & {\bf 46.3} & {\bf 47.1} & {\bf 46.3} & {\bf 17.3} & {\bf 17.2} & {\bf 17.6} & {\bf 18.6} & 14.6 \\ \hline
        \multirow{1}{*}{Pipe-ST}
            & Best system & -- & 37.9 & 40.3 & 39.2 & 17.6 & 17.2 & 16.7 & 18.5 & 14.0 \\ \hline
    \end{tabular}
    }
    \vspace{-2mm}
    \caption{Results of the end-to-end ST systems with pre-training}\label{tab:result_pretrain}
    \vspace{-5mm}
    \endgroup
\end{table*}

\section{Transfer learning for a very low-resource language speech translation}\label{ssec:transfer}
In this section, we evaluate generalization of multilingual ST models by performing transfer learning to a very low-resource ST task.
We used Mboshi-French corpus\footnote{\url{https://github.com/besacier/mboshi-french-parallel-corpus}} \cite{mboshi_french}, which contains 4.4-hours of spoken utterances and the corresponding Mboshi transcriptions and French translations.
Mboshi \cite{mboshi} is a Bantu C25 language spoken in Congo-Brazzaville and does not have standard orthography.
We sampled 100 utterances from the training set as the validation set, and report results on the dev set (514 utterances) as in \cite{tied_multitask,pretraining_st}.

We tried four different ways to transfer a non-Mboshi E2E-ST model to this task.
In the bilingual case, we used the bilingual ST model in Librispeech ((E-B-1) in Table \ref{tab:result_libri_bleu}) as seed, then fine-tuned on the Mboshi-French data.
In the multilingual case, we tried seeding with multilingual ST models in M2Ma (E-Ma-1), M2Mc (E-Mc-1), and O2M (E-O-1) settings.
All parameters including the output layer are transferred from pre-trained ST models and we do not include any characters in Mboshi transcriptions in the vocabularies.
Note that French references appear in the target side of all seed models during the pre-training stage.

Results are shown in Table \ref{tab:result_mboshi_st}.
Multilingual E2E-ST models are more effective than the bilingual one, and O2M showed the best performance among three models.
Although our transferred models underperformed \cite{pretraining_st}, it is worth mentioning that they used other English ASR data (Switchboard corpus) and initialized the decoder with the French ASR decoder.
Further improvements could be possible by leveraging Mboshi transcriptions, but we did not use any prior knowledge about Mboshi characters.
This is a desired scenario for endangered language documentation and quite useful for automatic word discovery \cite{tied_multitask,unwritten_asru2017,unsupervised_is18}.

\begin{table}[h]
    \centering
    \begingroup
    \begin{tabular}{l|l|c|c} \hline
     \multicolumn{2}{c|}{Seed} & \multirow{2}{*}{\shortstack{Multi-\\lingual}} & \multirow{2}{*}{BLEU} \\ \cline{1-2}
      Encoder & Decoder & &  \\ \hline
         En300h-ASR & French20h-ASR \cite{pretraining_st} & -- & {\bf 7.1} \\ \hline
         Libri-ST & Libri-ST & -- & 4.55 \\
         O2M-ST & O2M-ST & $\checkmark$ & {\bf 6.92} \\
         M2Ma-ST & M2Ma-ST & $\checkmark$ & 5.50 \\
         M2Mc-ST & M2Mc-ST & $\checkmark$ & 6.52 \\ \hline
      \end{tabular}
      \caption{Results of E2E-ST systems transferred from pre-trained E2E-ST models on a very low-resource corpus (Mboshi$\to$Fr, 4.4 hours). The former and latter part of hyphen represents \textit{data} and \textit{task} for pre-training, respectively (\textit{data}-\textit{task}). Note that all models do not use any transcriptions in Mboshi during pre-training nor adaptation stage.}\label{tab:result_mboshi_st}
      \vspace{-4mm}
    \endgroup
\end{table}

\section{Related work}\label{sec:related_work}
\subsubsection*{End-to-end speech translation}
\vspace{-1mm}
In \cite{google_fisher_st}, the E2E-ST model is simultaneously optimized with an auxiliary ASR task by sharing the whole encoder parameters.
Pre-training approaches from the ASR encoder \cite{pretraining_st} and MT decoder are also investigated in \cite{audiobook_st,google_tts_augmentation}.
\cite{google_tts_augmentation} proposed a data augmentation strategy, where weakly-supervised paired data is generated from monolingual source text data with text-to-speech (TTS) and MT systems (similar to back translation \cite{back_translation}) and speech data with a pipeline ST system (similar to knowledge distillation \cite{back_translation}).
\cite{tied_multitask} proposed an efficient framework to better leverage higher-level intermediate representations by jointly attending to speech encoder and transcription decoder states.
The most relevant work to ours is \cite{pretraining_st}, where well-trained ASR parameters from the other language are used to initialize ST models and improve the ST performance in low-resource scenarios.
Our work is distinct in that we focus on exploiting corpora in the multilingual setting and show that it outperforms the bilingual setting.
\vspace{-6mm}

\subsubsection*{Multilingual ASR}
\vspace{-1mm}
In the multilingual ASR study, the language-independent acoustic representations can be obtained by sharing parameters, and then adapted to low-resource languages \cite{language_independent_bnf,vu2014multilingual,martin_analysis}.
Recently, this approach is extended to end-to-end ASR paradigms: Connectionist Temporal Classification (CTC) \cite{dalmia2018sequence}, and attention-based encoder-decoder \cite{watanabe2017language,toshniwal2018multilingual,jj_slt18,inaguma2019transfer}.
Our work adopts this multilingual ASR in the pipeline system.
\vspace{-2mm}

\subsubsection*{Multilingual NMT}
\vspace{-1mm}
Crosslingual parameter sharing approaches are investigated by tying a part of parameters \cite{multi_nmt_multiway,dong2015multi,multi_nmt_interlingua}, and even all parameters with a shared vocabulary \cite{multi_nmt_google,multi_nmt_kit_iwslt16,multi_nmt_kit_iwslt17} among multiple languages.
Since the main drawback of the shared vocabulary is that the size of the vocabulary grows rapidly in proportion to the number of language pairs or the capacity per language shrinks when using BPE units \cite{sennrich2015neural}, fully character-level multilingual framework is proposed to overcome the issue to some extent \cite{fully_character}.
Our work is along with this trend of utilizing a universal translation model in one-to-many and many-to-many ST scenarios.




\section{Conclusion and future work}\label{sec:conclusion}
We performed multilingual training and end-to-end speech translation jointly, which has not yet been investigated before.
We proposed a universal sequence-to-sequence framework and it outperformed the bilingual end-to-end, and the gap between strong pipeline systems became smaller.
Its effectiveness was also confirmed by performing transfer learning to a very low-resource speech translation task.
To encourage further research in this topic, we will place our codes to the public project.
In future work, we will support more languages \cite{st_distillation,mustc} on our codebase and investigate multilingual training with non-related languages such as Chinese and Japanese.


\footnotesize
\bibliographystyle{IEEEbib}
\bibliography{refs}

\begin{thebibliography}{10}

\bibitem{ney1999speech}
Hermann Ney,
\newblock ``Speech translation: Coupling of recognition and translation,''
\newblock in {\em Proceedings of ICASSP}. IEEE, 1999, pp. 517--520.

\bibitem{punctuation_insertion}
Eunah Cho, Jan Niehues, and Alex Waibel,
\newblock ``{NMT}-based segmentation and punctuation insertion for real-time
  spoken language translation.,''
\newblock in {\em Proceedings of Interspeech}, 2017, pp. 2645--2649.

\bibitem{listen_and_translate}
Alexandre B{\'e}rard, Olivier Pietquin, Christophe Servan, and Laurent
  Besacier,
\newblock ``Listen and translate: A proof of concept for end-to-end
  speech-to-text translation,''
\newblock {\em arXiv preprint arXiv:1612.01744}, 2016.

\bibitem{google_fisher_st}
Ron~J Weiss, Jan Chorowski, Navdeep Jaitly, Yonghui Wu, and Zhifeng Chen,
\newblock ``Sequence-to-sequence models can directly translate foreign
  speech,''
\newblock in {\em Proceedings of Interspeech}, 2017, pp. 2625--2629.

\bibitem{audiobook_st}
Alexandre B{\'e}rard, Laurent Besacier, Ali~Can Kocabiyikoglu, and Olivier
  Pietquin,
\newblock ``End-to-end automatic speech translation of audiobooks,''
\newblock in {\em Proceedings of ICASSP}. IEEE, 2018, pp. 6224--6228.

\bibitem{acoustic_for_nmt}
Salil Deena, Raymond~WM Ng, Pranava Madhyastha, Lucia Specia, and Thomas Hain,
\newblock ``Exploring the use of acoustic embeddings in neural machine
  translation,''
\newblock in {\em Proceedings of ASRU}. IEEE, 2017, pp. 450--457.

\bibitem{st_distillation}
Yuchen Liu, Hao Xiong, Zhongjun He, Jiajun Zhang, Hua Wu, Haifeng Wang, and
  Chengqing Zong,
\newblock ``End-to-end speech translation with knowledge distillation,''
\newblock {\em arXiv preprint arXiv:1904.08075}, 2019.

\bibitem{google_tts_augmentation}
Ye~Jia, Melvin Johnson, Wolfgang Macherey, Ron~J Weiss, Yuan Cao, Chung-Cheng
  Chiu, Naveen Ari, Stella Laurenzo, and Yonghui Wu,
\newblock ``Leveraging weakly supervised data to improve end-to-end
  speech-to-text translation,''
\newblock in {\em Proceedings of ICASSP}. IEEE, 2019.

\bibitem{wit3}
Mauro Cettolo, Christian Girardi, and Marcello Federico,
\newblock ``Wit3: {W}eb inventory of transcribed and translated talks,''
\newblock in {\em Conference of European Association for Machine Translation},
  2012, pp. 261--268.

\bibitem{dessloch2018kit}
Florian Dessloch, Thanh-Le Ha, Markus M{\"u}ller, Jan Niehues, Thai~Son Nguyen,
  Ngoc-Quan Pham, Elizabeth Salesky, Matthias Sperber, Sebastian St{\"u}ker,
  Thomas Zenkel, et~al.,
\newblock ``Kit lecture translator: Multilingual speech translation with
  one-shot learning,''
\newblock in {\em Proceedings of the 27th International Conference on
  Computational Linguistics: System Demonstrations}, 2018, pp. 89--93.

\bibitem{watanabe2017language}
Shinji Watanabe, Takaaki Hori, and John~R Hershey,
\newblock ``Language independent end-to-end architecture for joint language
  identification and speech recognition,''
\newblock in {\em Proceedings of ASRU}. IEEE, 2017, pp. 265--271.

\bibitem{toshniwal2018multilingual}
Shubham Toshniwal, Tara~N Sainath, Ron~J Weiss, Bo~Li, Pedro Moreno, Eugene
  Weinstein, and Kanishka Rao,
\newblock ``Multilingual speech recognition with a single end-to-end model,''
\newblock in {\em Proceedings of ICASSP}. IEEE, 2018, pp. 4904--4908.

\bibitem{dalmia2018sequence}
Siddharth Dalmia, Ramon Sanabria, Florian Metze, and Alan~W Black,
\newblock ``Sequence-based multi-lingual low resource speech recognition,''
\newblock in {\em Proceedings of ICASSP}. IEEE, 2018, pp. 4909--4913.

\bibitem{jj_slt18}
Jaejin Cho, Murali~Karthick Baskar, Ruizhi Li, Matthew Wiesner, Sri~Harish
  Mallidi, Nelson Yalta, Martin Karafiat, Shinji Watanabe, and Takaaki Hori,
\newblock ``Multilingual sequence-to-sequence speech recognition: architecture,
  transfer learning, and language modeling,''
\newblock in {\em Proceedings of SLT}. IEEE, 2018, pp. 512--527.

\bibitem{inaguma2019transfer}
Hirofumi Inaguma, Jaejin Cho, Murali~Karthick Baskar, Tatsuya Kawahara, and
  Shinji Watanabe,
\newblock ``Transfer learning of language-independent end-to-end asr with
  language model fusion,''
\newblock in {\em Proceedings of ICASSP}. IEEE, 2019, pp. 6096--6100.

\bibitem{multi_nmt_google}
Melvin Johnson, Mike Schuster, Quoc~V Le, Maxim Krikun, Yonghui Wu, Zhifeng
  Chen, Nikhil Thorat, Fernanda Vi{\'e}gas, Martin Wattenberg, Greg Corrado,
  et~al.,
\newblock ``Google's multilingual neural machine translation system: enabling
  zero-shot translation,''
\newblock {\em Transactions of the Association for Computational Linguistics},
  2016.

\bibitem{multi_nmt_kit_iwslt16}
Thanh-Le Ha, Jan Niehues, and Alexander Waibel,
\newblock ``Toward multilingual neural machine translation with universal
  encoder and decoder,''
\newblock in {\em Proceedings of IWSLT}, 2016.

\bibitem{multi_nmt_kit_iwslt17}
Thanh-Le Ha, Jan Niehues, and Alexander Waibel,
\newblock ``Effective strategies in zero-shot neural machine translation,''
\newblock in {\em Proceedings of IWSLT}, 2017, pp. 105--112.

\bibitem{fisher_callHome}
Matt Post, Gaurav Kumar, Adam Lopez, Damianos Karakos, Chris Callison-Burch,
  and Sanjeev Khudanpur,
\newblock ``Improved speech-to-text translation with the {Fisher and Callhome
  Spanish--English} speech translation corpus,''
\newblock in {\em Proceedings of IWSLT}, 2013.

\bibitem{librispeech_french}
Ali~Can Kocabiyikoglu, Laurent Besacier, and Olivier Kraif,
\newblock ``Augmenting {Librispeech with French} translations: A multimodal
  corpus for direct speech translation evaluation,''
\newblock in {\em Proceedings of LREC}, 2018.

\bibitem{jan2018iwslt}
Niehues Jan, Roldano Cattoni, St{\"u}ker Sebastian, Mauro Cettolo, Marco
  Turchi, and Marcello Federico,
\newblock ``The iwslt 2018 evaluation campaign,''
\newblock in {\em Proceedings of IWSLT}, 2018, pp. 2--6.

\bibitem{mboshi_french}
Pierre Godard, Gilles Adda, Martine Adda-Decker, Juan Benjumea, Laurent
  Besacier, Jamison Cooper-Leavitt, Guy-No{\"e}l Kouarata, Lori Lamel,
  H{\'e}l{\`e}ne Maynard, Markus M{\"u}ller, et~al.,
\newblock ``A very low resource language speech corpus for computational
  language documentation experiments,''
\newblock {\em arXiv preprint arXiv:1710.03501}, 2017.

\bibitem{transformer}
Ashish Vaswani, Noam Shazeer, Niki Parmar, Jakob Uszkoreit, Llion Jones,
  Aidan~N. Gomez, Lukasz Kaiser, and Illia Polosukhin,
\newblock ``Attention is all you need,''
\newblock in {\em Processings of NIPS}, 2017.

\bibitem{attention_nmt_bahdanau}
Dzmitry Bahdanau, Kyunghyun Cho, and Yoshua Bengio,
\newblock ``Neural machine translation by jointly learning to align and
  translate,''
\newblock in {\em Proceedings of ICLR}, 2015.

\bibitem{las}
William Chan, Navdeep Jaitly, Quoc Le, and Oriol Vinyals,
\newblock ``Listen, attend and spell: A neural network for large vocabulary
  conversational speech recognition,''
\newblock in {\em Proceedings of ICASSP}. IEEE, 2016, pp. 4960--4964.

\bibitem{sennrich2015neural}
Rico Sennrich, Barry Haddow, and Alexandra Birch,
\newblock ``Neural machine translation of rare words with subword units,''
\newblock in {\em Proceedings of ACL}, 2016, pp. 1715--1725.

\bibitem{hori2017joint}
Takaaki Hori, Shinji Watanabe, and John Hershey,
\newblock ``Joint {CTC}/attention decoding for end-to-end speech recognition,''
\newblock in {\em Proceedings of ACL}, 2017, pp. 518--529.

\bibitem{hybrid_ctc_attention}
Shinji Watanabe, Takaaki Hori, Suyoun Kim, John~R Hershey, and Tomoki Hayashi,
\newblock ``Hybrid {CTC}/attention architecture for end-to-end speech
  recognition,''
\newblock {\em IEEE Journal of Selected Topics in Signal Processing}, vol. 11,
  no. 8, pp. 1240--1253, 2017.

\bibitem{ctc_graves}
Alex Graves, Santiago Fern{\'a}ndez, Faustino Gomez, and J{\"u}rgen
  Schmidhuber,
\newblock ``Connectionist temporal classification: labelling unsegmented
  sequence data with recurrent neural networks,''
\newblock in {\em Proceedings of ICML}, 2006, pp. 369--376.

\bibitem{attention_nips2015}
Jan~K Chorowski, Dzmitry Bahdanau, Dmitriy Serdyuk, Kyunghyun Cho, and Yoshua
  Bengio,
\newblock ``Attention-based models for speech recognition,''
\newblock in {\em Proceedings of NIPS}, 2015, pp. 577--585.

\bibitem{rnnlm}
Tom{\'a}{\v{s}} Mikolov, Martin Karafi{\'a}t, Luk{\'a}{\v{s}} Burget, Jan
  {\v{C}}ernock{\`y}, and Sanjeev Khudanpur,
\newblock ``Recurrent neural network based language model,''
\newblock in {\em Proceedings of Interspeech}, 2010.

\bibitem{speed_perturbation}
Tom Ko, Vijayaditya Peddinti, Daniel Povey, and Sanjeev Khudanpur,
\newblock ``Audio augmentation for speech recognition,''
\newblock in {\em Proceedings of Interspeech}, 2015, pp. 3586--3589.

\bibitem{multi_nmt_multiway}
Orhan Firat, Kyunghyun Cho, and Yoshua Bengio,
\newblock ``Multi-way, multilingual neural machine translation with a shared
  attention mechanism,''
\newblock in {\em Proceedings of NAACL-HLT}, 2016, pp. 866--875.

\bibitem{mustc}
Mattia~A Di~Gangi, Roldano Cattoni, Luisa Bentivogli, Matteo Negri, and Marco
  Turchi,
\newblock ``{MuST-C}: a multilingual speech translation corpus,''
\newblock in {\em Proceedings of the NAACL-HLT}, 2019, pp. 2012--2017.

\bibitem{some_insights}
Gaurav Kumar, Matt Post, Daniel Povey, and Sanjeev Khudanpur,
\newblock ``Some insights from translating conversational telephone speech,''
\newblock in {\em Proceedings of ICASSP}. IEEE, 2014, pp. 3231--3235.

\bibitem{librispeech}
Vassil Panayotov, Guoguo Chen, Daniel Povey, and Sanjeev Khudanpur,
\newblock ``Librispeech: an {ASR} corpus based on public domain audio books,''
\newblock in {\em Proceedings of ICASSP}. IEEE, 2015, pp. 5206--5210.

\bibitem{kaldi}
Daniel Povey, Arnab Ghoshal, Gilles Boulianne, Lukas Burget, Ondrej Glembek,
  Nagendra Goel, Mirko Hannemann, Petr Motlicek, Yanmin Qian, Petr Schwarz,
  et~al.,
\newblock ``The {Kaldi} speech recognition toolkit,''
\newblock in {\em Proceedings of ASRU}. IEEE, 2011.

\bibitem{iwslt18_data_cleaning}
Mattia~Antonino Di~Gangi, Roberto Dess{\`\i}, Roldano Cattoni, Matteo Negri,
  and Marco Turchi,
\newblock ``Fine-tuning on clean data for end-to-end speech translation:
  {FBK}@{IWSLT} 2018,''
\newblock in {\em Proceedings of IWSLT}, 2018, pp. 147--152.

\bibitem{disfluency_st}
Elizabeth Salesky, Susanne Burger, Jan Niehues, and Alex Waibel,
\newblock ``Towards fluent translations from disfluent speech,''
\newblock in {\em Proceedings of SLT}. IEEE, 2018, pp. 921--926.

\bibitem{bleu}
Kishore Papineni, Salim Roukos, Todd Ward, and Wei-Jing Zhu,
\newblock ``Bleu: a method for automatic evaluation of machine translation,''
\newblock in {\em Proceedings of ACL}, 2002, pp. 311--318.

\bibitem{vgg}
Karen Simonyan and Andrew Zisserman,
\newblock ``Very deep convolutional networks for large-scale image
  recognition,''
\newblock in {\em Proceedings of ICLR}, 2015.

\bibitem{lstm}
Sepp Hochreiter and J{\"u}rgen Schmidhuber,
\newblock ``Long short-term memory,''
\newblock {\em Neural Computation}, vol. 9, no. 8, pp. 1735--1780, 1997.

\bibitem{adadelta}
Matthew~D Zeiler,
\newblock ``Adadelta: an adaptive learning rate method,''
\newblock {\em arXiv preprint arXiv:1212.5701}, 2012.

\bibitem{adam}
Diederik Kingma and Jimmy Ba,
\newblock ``Adam: A method for stochastic optimization,''
\newblock {\em arXiv preprint arXiv:1412.6980}, 2014.

\bibitem{zaremba2014recurrent}
Wojciech Zaremba, Ilya Sutskever, and Oriol Vinyals,
\newblock ``Recurrent neural network regularization,''
\newblock {\em arXiv preprint arXiv:1409.2329}, 2014.

\bibitem{label_smoothing}
Christian Szegedy, Vincent Vanhoucke, Sergey Ioffe, Jon Shlens, and Zbigniew
  Wojna,
\newblock ``Rethinking the inception architecture for computer vision,''
\newblock in {\em Proceedings of CVPR}, 2016, pp. 2818--2826.

\bibitem{scheduled_sampling}
Samy Bengio, Oriol Vinyals, Navdeep Jaitly, and Noam Shazeer,
\newblock ``Scheduled sampling for sequence prediction with recurrent neural
  networks,''
\newblock in {\em Proceedings of NIPS}, 2015, pp. 1171--1179.

\bibitem{pretraining_st}
Sameer Bansal, Herman Kamper, Karen Livescu, Adam Lopez, and Sharon Goldwater,
\newblock ``Pre-training on high-resource speech recognition improves
  low-resource speech-to-text translation,''
\newblock in {\em Proceedings of NAACL-HLT}, 2019, pp. 58--68.

\bibitem{mboshi}
Annie Rialland, Martine Adda-Decker, Guy-No{\"e}l Kouarata, Gilles Adda,
  Laurent Besacier, Lori Lamel, Elodie Gauthier, Pierre Godard, and Jamison
  Cooper-Leavitt,
\newblock ``Parallel corpora in {Mboshi (Bantu C25, Congo-Brazzaville}),''
\newblock in {\em Proceedings of LREC}, 2018, pp. 4272--4276.

\bibitem{tied_multitask}
Antonios Anastasopoulos and David Chiang,
\newblock ``Tied multitask learning for neural speech translation,''
\newblock in {\em Proceedings of NAACL-HLT}, 2018, pp. 82--91.

\bibitem{unwritten_asru2017}
Marcely~Zanon Boito, Alexandre B{\'e}rard, Aline Villavicencio, and Laurent
  Besacier,
\newblock ``Unwritten languages demand attention too! word discovery with
  encoder-decoder models,''
\newblock in {\em Proceedings of ASRU}. IEEE, 2017, pp. 458--465.

\bibitem{unsupervised_is18}
Pierre Godard, Marcely Zanon-Boito, Lucas Ondel, Alexandre Berard,
  Fran{\c{c}}ois Yvon, Aline Villavicencio, and Laurent Besacier,
\newblock ``Unsupervised word segmentation from speech with attention,''
\newblock in {\em Proceedings of Interspeech}, 2018, pp. 2678--2782.

\bibitem{back_translation}
Rico Sennrich, Barry Haddow, and Alexandra Birch,
\newblock ``Improving neural machine translation models with monolingual
  data,''
\newblock in {\em Proceedings of ACL}, 2016, pp. 86--96.

\bibitem{language_independent_bnf}
Karel Vesel{\`y}, Martin Karafi{\'a}t, Franti{\v{s}}ek Gr{\'e}zl, Milo{\v{s}}
  Janda, and Ekaterina Egorova,
\newblock ``The language-independent bottleneck features,''
\newblock in {\em Proceedings of SLT}. IEEE, 2012, pp. 336--341.

\bibitem{vu2014multilingual}
Ngoc~Thang Vu, David Imseng, Daniel Povey, Petr Motlicek, Tanja Schultz, and
  Herv{\'e} Bourlard,
\newblock ``Multilingual deep neural network based acoustic modeling for rapid
  language adaptation,''
\newblock in {\em Proceeding of ICASSP}. IEEE, 2014, pp. 7639--7643.

\bibitem{martin_analysis}
Martin Karafi{\'a}t, Murali~Karthick Baskar, Karel Vesel{\`y}, Franti{\v{s}}ek
  Gr{\'e}zl, Luk{\'a}{\v{s}} Burget, et~al.,
\newblock ``Analysis of multilingual blstm acoustic model on low and high
  resource languages,''
\newblock in {\em Proceeding of ICASSP}. IEEE, 2018, pp. 5789--5793.

\bibitem{dong2015multi}
Daxiang Dong, Hua Wu, Wei He, Dianhai Yu, and Haifeng Wang,
\newblock ``Multi-task learning for multiple language translation,''
\newblock in {\em Proceedings of ACL}, 2015, pp. 1723--1732.

\bibitem{multi_nmt_interlingua}
Yichao Lu, Phillip Keung, Faisal Ladhak, Vikas Bhardwaj, Shaonan Zhang, and
  Jason Sun,
\newblock ``A neural interlingua for multilingual machine translation,''
\newblock in {\em Proceedings of the Third Conference on Machine Translation
  (WMT)}, 2018, pp. 84--92.

\bibitem{fully_character}
Jason Lee, Kyunghyun Cho, and Thomas Hofmann,
\newblock ``Fully character-level neural machine translation without explicit
  segmentation,''
\newblock {\em Transactions of the Association for Computational Linguistics},
  2017.

\end{thebibliography}

\end{document}